\def\year{2019}\relax
\documentclass[letterpaper]{article} %DO NOT CHANGE THIS
\usepackage{aaai19}  %Required
\usepackage{times}  %Required
\usepackage{helvet}  %Required
\usepackage{courier}  %Required
\usepackage{url}  %Required
\usepackage{graphicx}  %Required
\usepackage{graphicx}
\usepackage{amsmath,amssymb} % define this before the line numbering.
\usepackage{color}
\usepackage{tikz}
\usepackage{color}
\usepackage{pgfpages}
\usepackage{pgfplots}
\usepackage{algorithm}
\usepackage{lipsum,multicol}
\usepackage{algorithmic}
\usepackage{enumitem}
\begin{document}
\title{Transfer Incremental Learning using Data Augmentation}
\author{Ghouthi Boukli Hacene, Vincent Gripon, Nicolas Farrugia, Matthieu Arzel, Michel Jezequel\\
\\
IMT Atlantique, Brest, France\\
\\
}
\maketitle
\begin{abstract}
Deep learning-based methods have reached state of the art performances, relying on large quantity of available data and computational power. Such methods still remain highly inappropriate when facing a major open machine learning problem, which consists of learning incrementally new classes and examples over time.
Combining the outstanding performances of Deep Neural Networks (DNNs) with the flexibility of incremental learning techniques
is a promising venue of research.
In this contribution, we introduce Transfer Incremental Learning using Data Augmentation (TILDA). TILDA is based on pre-trained DNNs as feature extractor, robust selection of feature vectors in subspaces using a nearest-class-mean based technique, majority votes and data augmentation at both the training and the prediction stages.
Experiments on challenging vision datasets demonstrate the ability of
the proposed method for low complexity incremental learning, while achieving
significantly better accuracy than existing incremental counterparts.
\end{abstract}

\section{Introduction}

Humans have the ability to incrementally learn new pieces of information through time, building over previously acquired knowledge. This process is most of the time nondestructive, and results in what is often referred to as ``curriculum learning'' in the literature~\cite{bengio2009curriculum}. On the contrary, it has been known for decades that neural networks learning procedures, despite the fact they originally were proposed as a simplifying model for brain mechanisms, suffer from ``catastrophic forgetting''~\cite{kasabov2013evolving,french1999catastrophic}, or the fact that previously learned knowledge is destroyed when learning new one.

During last years, deep learning has become the golden standard in many supervised learning challenges, especially in the field of computer vision~\cite{iandola2016squeezenet,DBLP:journals/corr/SimonyanZ14a,szegedy2015rethinking}. Deep Learning relies on the use of a large number of trainable parameters, that are carefully adjusted using stochastic gradient descent based algorithms. Learning novel data using the same set of parameters inevitably leads to the loss of the previously acquired knowledge. This is why many techniques have proposed to learn
distinct deep learning systems over the course of time, letting another algorithm decide which one to use at prediction stage~\cite{girshick2014rich,pan2010survey}. Such methods can quickly result in very complex systems, that are likely to fail in adversarial conditions.

Formally, an incremental learning approach would satisfy the following criteria~\cite{rebuffi-cvpr2017}:
\begin{enumerate}

\item An ability to learn data using one (or a few) example(s) at a time, in any order, without requiring to reconsider or store previous ones.
\item An ability to sustain a classification accuracy comparable to state-of-art methods while traversing successive incremental learning stages, thus avoiding catastrophic forgetting.
%\item An ability to classify at any instant with an accuracy comparable to state-of-art methods, thus avoid catastrophic forgetting.
\item Low computation and memory footprints, during training and classifying phases, that should remain sublinear in both of number of examples and their dimension.

\end{enumerate}

Satisfying these three criteria while keeping competitive accuracy of the proposed systems has remained a key open challenge.

A promising venue of research lies in ``transfer learning'' methods~\cite{girshick2014rich}, that make use of very efficient pre-trained deep neural networks previously obtained using huge datasets of signals related to the tasks at hand. As a result, very high quality feature vectors can be used to feed the incremental learning techniques, which can then achieve reasonable performances despite using simplistic
mechanisms~\cite{pan2010survey}.

In this paper, we introduce Transfer Increment Learning with Data Augmentation (TILDA), an incremental learning method that provides a) a robust selection of feature vectors in subspaces, and b) prediction procedures making use of data-augmentation. We stress the method using challenging vision datasets, namely CIFAR10, CIFAR100 and ImageNet LSVRC 2012. As a result the proposed method allows us to:
\begin{itemize}
\item Perform incremental learning following the above-mentioned definition,
\item Approach state-of-the-art performances on vision datasets,
\item Reduce memory usage and computation time by several order of magnitude compared to other incremental approaches.
\end{itemize}
%[mettre la remarque sur learning on chip dans la conclusion plutôt,
%comme ouverture de ton travail]

%The outline of the paper is as follows. We present the previous works in Section~\ref{previous work}. In Section~\ref{propose} we introduce the TILDA method. In Section~\ref{experiments}, we discuss results on challenging datasets and compare them with other methods. And finally, we conclude in Section~\ref{conclusion} and discuss future work. 

%%%%%%%%%%%%%%%%%%%%%%%%%%%%%%%%%%%%%%%%%%%%%%%%%%%%%%%%%%%%%%%%%%%%%%%%%%%%%%%%%%%%%%%%%%%%%%%%%%%%%%%%%%%%%%%%%%%%%%%%%%%%%%%%%%%%%%%%%%%%%%%%%%%%%%%%%%%ù

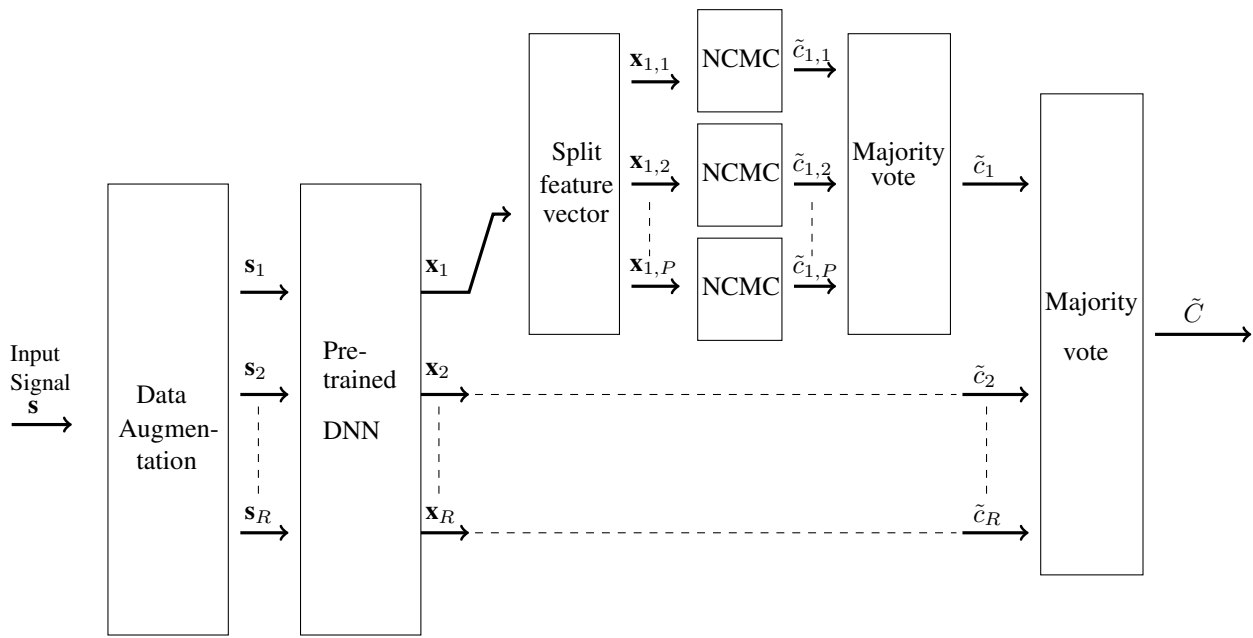
\begin{figure*}
\begin{tikzpicture}[scale=0.8]

\node[text width=1cm] at (0,-0.6) 
    {\footnotesize Input \\Signal};
\node[text width=1cm] at (0.3,-1.2) 
    {$\textbf{s}$};
\draw[->,very thick] (-0.6,-1.5) -- (0.4,-1.5);
\draw (1,-5) --  (3,-5) -- (3,2.5) -- (1,2.5) -- cycle;
\node[text width=1cm] at (2.1,-1) 
    {Data};
\node[text width=1cm] at (1.8,-1.8) 
    {Augmen-};
\node[text width=1cm] at (2.1,-2.1) 
    {tation};

\draw[->,very thick] (3.2,.7) -- (4,0.7);
\draw[->,very thick] (3.2,-1) -- (4,-1);
\draw[->,very thick] (3.2,-3.3) -- (4,-3.3);
\draw[dashed] (3.5,-1.2) -- (3.5,-2.7);
\node[text width=1cm] at (3.9,1.1) 
    {$\textbf{s}_1$};
\node[text width=1cm] at (3.9,-0.6) 
    {$\textbf{s}_2$};
\node[text width=1cm] at (3.9,-3) 
    {$\textbf{s}_R$};

 \draw (4.2,-5)--(6.2,-5)--(6.2,2.5)--(4.2,2.5)-- cycle;   
%\fill[orange] (4.2,-5) --  (4.7,-4.5) -- (4.7,2.5) -- (4.2,2) -- cycle;
%\fill[orange] (4.8,-5) --  (5.3,-4.5) -- (5.3,2.5) -- (4.8,2) -- cycle;
%\fill[orange] (5.5,-5) --  (6,-4.5) -- (6,2.5) -- (5.5,2) -- cycle;

\node[text width=1cm] at (5.2,-0.5) 
    {Pre-trained};
\node[text width=1cm] at (5.2,-1.6) 
    {DNN};
    
\draw[->,very thick] (6.2,.7) -- (7,.7) -- (7.4,2)--(7.7,2);
\draw[->,very thick] (6.2,-1) -- (7,-1);
\draw[->,very thick] (6.2,-3.3) -- (7,-3.3);
\draw[dashed] (6.5,-1.2) -- (6.5,-2.7);
\node[text width=1cm] at (6.9,1.1) 
    {$\textbf{x}_1$};
\node[text width=1cm] at (6.9,-0.6) 
    {$\textbf{x}_2$};
\node[text width=1cm] at (6.9,-3) 
    {$\textbf{x}_R$};

\draw (8,0) --  (9.5,0) -- (9.5,5) -- (8,5) -- cycle;
\node[text width=1cm] at (9,3) 
    {Split};
\node[text width=1cm] at (8.8,2.5) 
    {feature};
\node[text width=1cm] at (8.85,2) 
    {vector};

\draw[->,very thick] (9.7,4.2) -- (10.5,4.2);
\draw[->,very thick] (9.7,2.5) -- (10.5,2.5);
\draw[->,very thick] (9.7,.8) -- (10.5,.8);
\draw[dashed] (10,2.2) -- (10,1.3);
\node[text width=1cm] at (10.3,4.5) 
    {$\textbf{x}_{1,1}$};
\node[text width=1cm] at (10.3,2.8) 
    {$\textbf{x}_{1,2}$};
\node[text width=1cm] at (10.3,1.1) 
    {$\textbf{x}_{1,P}$};

\draw (10.8,5.4) --  (12.2,5.4) -- (12.2,3.7) -- (10.8,3.7) -- cycle;
\draw (10.8,3.5) --  (12.2,3.5) -- (12.2,1.8) -- (10.8,1.8) -- cycle;
\draw (10.8,1.6) --  (12.2,1.6) -- (12.2,-.1) -- (10.8,-.1) -- cycle;
\node[text width=1cm] at (11.5,4.6) 
    {NCMC};
\node[text width=1cm] at (11.5,2.7) 
    {NCMC};
\node[text width=1cm] at (11.5,.8) 
    {NCMC};
    
\draw[->,very thick] (12.4,4.4) -- (13.1,4.4);
\draw[->,very thick] (12.4,2.5) -- (13.1,2.5);
\draw[->,very thick] (12.4,.8) -- (13.1,.8);
\draw[dashed] (12.7,2.2) -- (12.7,1.3);
\node[text width=1cm] at (13,4.7) 
    {$\tilde{c}_{1,1}$};
\node[text width=1cm] at (13,2.8) 
    {$\tilde{c}_{1,2}$};
\node[text width=1cm] at (13,1.1) 
    {$\tilde{c}_{1,P}$};

\draw (13.3,0) --  (15,0) -- (15,5) -- (13.3,5) -- cycle;
\node[text width=1cm] at (14,3) 
    {Majority};
\node[text width=1cm] at (14.3,2.6) 
    {vote};
    
\draw[->,very thick] (15.2,2.5) -- (16.3,2.5);
\node[text width=1cm] at (16,2.8) 
    {$\tilde{c}_{1}$};
    
\draw[dashed] (7.1,-1) -- (15.1,-1);
\draw[->,very thick] (15.2,-1) -- (16.3,-1);
\node[text width=1cm] at (16,-.7) 
    {$\tilde{c}_{2}$};
    
\draw[dashed] (7.1,-3.3) -- (15.1,-3.3);
\draw[->,very thick] (15.2,-3.3) -- (16.3,-3.3);
\node[text width=1cm] at (16,-3) 
    {$\tilde{c}_{R}$};
    
\draw (16.5,-4) --  (18.2,-4) -- (18.2,4) -- (16.5,4) -- cycle;

\draw[dashed] (15.6,-1.2) -- (15.6,-2.7);

\node[text width=1cm] at (17.2,.5) 
    {Majority};
\node[text width=1cm] at (17.5,-.3) 
    {vote};
\draw[->,very thick] (18.4,0) -- (20,0);
    \node[text width=1cm] at (19.5,.4) 
    {$\tilde{C}$};
\end{tikzpicture}
\caption{Overview of the proposed method. Given an input signal $\textbf{s}$, we first use data augmentation to generate a multiple version of the input signal $\left(\textbf{s}_r\right)_{1\leq r \leq R}$. Then we use a pre-trained DNN for feature extraction and obtain the corresponding feature vectors $\left(\textbf{x}_r\right)_{1\leq r \leq R}$. Subsequently, we split each feature vector $\textbf{x}_r$ into $P$ equal parts $\left(\textbf{x}_{r,p}\right)_{1\leq p \leq P}$, and classify each part $\textbf{x}_{r,p}$ using a NCM-inspired classifiers (NCMC) containing anchor vectors $\left(\textbf{Y}_{c,p}\right)_{1\leq c \leq C}$. We obtain a class for each part $c_{r,p}$ and do a majority vote to get the class of $\textbf{x}_r$. Finally, a second majority vote is done thanks to the obtained classes $\left(c_r\right)_{1\leq r \leq R}$ of all generated signals to get assigned class $\tilde{C}$ to the original input signal $s$.}  
\label{fig:Proposed Method Figure}
\end{figure*}

\section{Related Work}
\label{previous work}
%``Incremental learning'' term addresses three different concepts ~\cite{zhou2002hybrid}: example-incremental learning~\cite{iSVM,poggio2001incremental}, class-incremental learning~\cite{Learn++,tweet}, and attribute-incremental learning.

There has been interests in incremental learning for a long time~\cite{schlimmer1986case,thrun1996learning,zhou2002hybrid}. For example, methods have been proposed~\cite{iSVM,poggio2001incremental,zheng2013online} to address this problem with the aim at bounding memory footprint (c.f. criterion 3.). These approaches perform learning one subset at a time using Support Vector Machines (SVMs). More precisely, a new SVM is trained for each batch of new data, exploiting previous support vectors. Since the latter are not conveying the full extent of previous data, the newly trained SVM suffers from \emph{catastrophic forgetting}~\cite{kasabov2013evolving,french1999catastrophic}, and thus violate criterion 2 defined in the introduction.

Another incremental learning algorithm, called ``Learn++'' was introduced~\cite{learn++,muhlbaier2009learn}. This algorithm adds weak one-vs-all classifiers to accommodate new classes. Therefore, it may result in an excessive  computational complexity and memory usage, disobeying criterion 3. It also needs training data for all classes to occur repeatedly, which contradicts criterion 1.% This method is also used to add the incremental learning capability to SVMs, by using an ensemble of SVMs trained with Learn++ called ``SVMLearn++''~\cite{erdem2005ensemble}. This consists of using the learn++ algorithm with an SVM classifier. Despite the fact that SVMLearn++ showed a promising results on biological datasets~\cite{molina2014incremental}, this method still needs to train new SVM for each new data, and suffers from \emph{catastrophic forgetting}.   

%During the last decade, CNNs have became quickly the state-of-art~\cite{DBLP:journals/corr/HongYKH15,pan2010survey,krizhevsky2012imagenet} in numerous domains, which has made the previous incremental methods become outdated, because they can not be used to learn huge datasets as ImageNet, and the obtained accuracy by these methods and CNNs are no longer comparable. Based on this, new incremental methods have been proposed based on CNNs as features extractors~\cite{rebuffi-cvpr2017,hacenebudget,haceneincremental,shmelkov2017incremental}, and then the classification process is done on the obtained feature vectors.

Research showed also the possibility for the sequential learning of data~\cite{pentina2015curriculum}. however, this requires to choose a correct ordering of the whole dataset, which does not fulfil criterion 1. In~\cite{mensink2013distance}, the authors proposed to use a pre-trained and unchanged DNN as feature extractor followed by the Nearest Class Mean classifier (NCM). NCM summarises each class using the average feature vector of all examples observed for the class so far. Classification processes by assigning the class of the most similar average vector using a metric that can be learned from data. Compared to other parametric classifiers~\cite{mensink2012metric,mensink2013distance,ristin2014incremental}, NCM showed better performances in incremental learning scenarios. However, NCM gives a lower accuracy than state-of-art methods even when it uses all the dataset, hence does not fulfil criterion 2.

In~\cite{hacenebudget}, a quite different incremental method called Budget Restricted Incremental Learning (BRIL) was proposed. BRIL combines ``transfer learning''~\cite{girshick2014rich,pan2010survey} with binary associative memories. A pre-trained DNN is used as feature extractor, as mentioned in~\cite{mensink2013distance}, while binary associative memories act as a classifier. A product random sampling is performed as an intermediate between the pre-trained DNN and the classifier. Despite being compliant with criteria 1 and 3, the accuracy remains significantly lower than existing counterparts, which violates criterion 2.

  Kuzborskij et al~\cite{kuzborskij2013n} showed that new classes can be added to a multi-class classifier with limited impact on  accuracy when the classifiers can be retrained from at least a small amount of data belonging to all classes. Using this, in~\cite{rebuffi-cvpr2017} the authors proposed an incremental learning method called ``Incremental Classifier and Representation Learning'' (iCaRL), based on a trainable DNN feature extractor, followed by a single classification layer. The classification process is inspired by NCM: it computes the mean of feature vectors for each class, and assign the label of the nearest prototype. However, memory usage can easily increase, especially when the dataset is made of high resolution images such as ImageNet, which may violate criterion 3. Moreover, the iCaRL method, when trained on data streams containing only few classes at a time, provides low accuracy as shown in~\cite{rebuffi-cvpr2017}, hence iCaRL does not respect criterion 2. To reach good performances and a comparable accuracy to state-of-art methods, iCaRL thus needs to be trained over batches of data containing a large part of the dataset, which does not correspond to an incremental learning scenario and infringes~1.

In this paper, we introduce TILDA that builds upon previously proposed work, attempting to cover all 3 criteria for efficient incremental learning. 
As in iCaRL and BRIL, TILDA uses a pre-trained DNN as feature extractor. TILDA also uses an NCM-inspired classifier over the feature vectors obtained from by the pre-trained DNN. Data augmentation is performed on both training and classification datasets, aiming to improve accuracy. Consequently, there is no need to retrain the system with previous data, nor to perform computationally intensive processing when new data comes in. In addition, learning new data does not damage previously learned information.

\section{Proposed Method}
\label{propose}
In this section, we describe the TILDA method. We start by giving a high level overview of the process, and then we explain the details.

\subsection{Overview of the Proposed Method}
\label{overview}

TILDA is built upon four main steps: 1) a pre-trained DNN to perform feature extraction, 2) a technique to project features into low dimensional subspaces, 3) an assembly of NCM-inspired Classifiers (NCMC) applied independently in each subspace (c.f. Figure~\ref{fig:Proposed Method Figure}) and 4) a data augmentation inspired scheme to increase accuracy of the classifying process. We develop these steps in the following paragraphs.

The first step consists of using the internal layers of a pre-trained DNN~\cite{krizhevsky2012imagenet} as a generic feature extractor on which subsequent learning is performed. This process has become increasingly popular in the past few years and is often referred to as ``Transfer Learning''~\cite{Oquab_2014_CVPR}. The aim is to transfer acquired knowledge on a dataset to another related problem~\cite{pan2010survey}. 

In the following step we project feature vectors into multiple low dimensional subspaces. %% Because of criterion 3, we choose to implement product random sampling for its light computational footprint. More precisely, we split features into multiple subvectors, which are quantized independently from each others using random selection of anchor subvectors in the dataset.
More precisely, we split feature vectors into $P$ subvectors. For each class and each subspace, we produce $k$ anchor vectors conveying robust statistical properties about corresponding feature subvectors.

Then, in each subspace anchor vectors are exploited to perform weak classification of the input data. We use here a NCM inspired method. A majority vote is then performed to obtain an aggregate decision.

Finally, we perform data augmentation on the input signals, be them training or testing inputs, thus obtaining multiple decisions for each input data as well as more robust classifiers in each subspace. A second majority vote is performed using these decisions to generate a global prediction.

\subsection{Details of the Proposed Method}
\label{improvement}
\subsubsection{Pre-Trained Deep Neural Networks} \label{ssec:num1}

To obtain features from an input signal, TILDA relies on using DNNs that are pre-trained on a large number of examples. Consequently, using the pre-trained inner layers of the DNN acts as a generic feature extractor~\cite{Oquab_2014_CVPR,DBLP:journals/corr/HongYKH15,pan2010survey}. %% In this paper, our experiments focus on the use of Inception V3~\cite{szegedy2015rethinking}, that is trained on 1000 classes from ImageNet\footnote{ImageNet challenge url: \url{http://image-net.org/challenges/LSVRC/2012/index}}.
As a matter of fact, inner layers of a deep DNN offer a good generic description of an input image, even when it does not belong to the learning domain~\cite{Oquab_2014_CVPR}.

Using ``Transfer Learning'' ideas, we are not interested in this work in the network's architecture details, as we simply use the appropriate layers to extract features from a given input. %% Here, we take outputs of the last pooling layer\footnote{\url{https://github.com/Hvass-Labs/TensorFlow-Tutorials/blob/master/08_Transfer_Learning.ipynb}}. 

In the remainder of this paper, we denote by $\textbf{s}^m$ the $m$-th input training signal and by $\textbf{x}^m$ its corresponding feature vector, where $1 \leq m \leq M$ and $M$ is the total number of training signals.

\subsubsection{Projection to Low Dimensional Subspaces}

%% \subsubsection{Product Random Sampling} \label{ssec:num2}

Feature extraction allows us to consider the feature vector $\textbf{x}^m$ instead of the input signal $\textbf{s}^m$. Formally, let us denote $\textbf{x}_c^m$ the fact that feature vector $\textbf{x}^m$ belongs to class $c$. We split each $\textbf{x}_c^m$ into $P$ parts, denoted $\left(\textbf{x}_{c,p}^m\right)_{1\leq p \leq P}$. 
For each class and each subspace, we create $k$ anchor vectors initialised with 0s, each of them associated with a counter, also initialised by 0. Considering the $p$-th subspace and the $c$-th class, we denote by $Y_{c,p}=[\textbf{y}_{c,p,1},..., \textbf{y}_{c,p,k}]$ the corresponding anchor vectors and $N_{c,p}=[n_{c,p,1},\dots, n_{c,p,k}]$ their associated counters.

For each $c$ and $p$, we aim at using the corresponding anchor vectors as centroids of a clustering of $\{\textbf{x}_{c,p}^m\}$. To this end, at each step of the training process, we ensure that each anchor vector is a centroid of a clustering of already processed input subvectors, and the associated counter accounts for the cardinality of the corresponding cluster.

Then, each time an input training vector is processed, we identify an anchor vector to be updated. The update simply consists of computing a new anchor vector obtained as a barycenter of the old one with weight given by its counter and the input subvector with weight 1, then incrementing the counter. This procedure is detailed in~Algorithm~\ref{Algo:one}. Namely, rather than simply associating the new subvector with the closest anchor vector, what would inevitably lead to unbalanced counters and thus poor performance in prediction, we prefer to take into account counters while performing this association. More precisely, we linearly penalize anchor vectors that are already made of the combination of many subvectors. Note that when two or more anchor vectors gives the same results (distances multiplied by counters), we choose uniformly at random one of these anchor vectors.

%% $\textbf{C}_{p,t}=[c_{t,1},...,c_{t,k}]$ where $c_{t,1}=c_{t,k}=c_t$ . We repeat this process for each new class, allowing incremental learning, and avoiding catastrophic forgetting because we do not modify the prior knowledge (c.f. Algorithm~\ref{Algo:one}).

\begin{algorithm}
\caption{Incremental Learning of Anchor Subvectors}
 \textbf{Input}: streaming feature vector $\textbf{x}_c^{m}$\\

\begin{algorithmic}

\FOR{$p:=1$ to $P$}
\FOR{$i:=1$ to $k$}
\STATE $d_i=\| \textbf{x}_{p} - \textbf{y}_{c,p,i}\|_2$
\STATE $R_i=d_i n_{c,p,i}$

\ENDFOR
\STATE $\tilde{k}=\displaystyle{\arg\min_{i}R_i}$
\STATE $\textbf{y}_{c,p,\tilde{k}} \leftarrow \textbf{y}_{c,p,\tilde{k}} n_{c,p,\tilde{k}} + \textbf{x}_{c,p}^{m}$
\STATE $n_{c,p,\tilde{k}} \leftarrow n_{c,p,\tilde{k}} + 1$
\STATE $\textbf{y}_{c,p,\tilde{k}} \leftarrow \textbf{y}_{c,p,\tilde{k}}/n_{c,p,\tilde{k}}$
\ENDFOR

\end{algorithmic}
%% \\{\algorithmicindent} \textbf{Output} : $\left(Y_p\right)_{1\leq p \leq P}$, $\left(C_p\right)_{1\leq p \leq P}$. \\
\label{Algo:one}
\end{algorithm}

Note that the learning process is independent on the order of streaming data, and is performed one example at a time, thus enforcing criterion 1 of incremental learning methods described in the introduction.
%% Note that the learning process weakly depends on the order in which input data is streamed. 
%% We assume that the model start learning from scratch, and at the beginning $Y_p=C_p=\oldemptyset$.

\subsubsection{Aggregation of subspaces weak classifiers}

At prediction stage, consider an input signal $\textbf{s}$ and the associated feature vector $\textbf{x}$. We split $\textbf{x}$ into the corresponding $P$ parts and obtain $\left(\textbf{x}_p\right)_{1\leq p \leq P}$. We compute Euclidean distances between each $\textbf{x}_p$ and all anchor subvectors $\textbf{y}_{c,p,i}$ for which the counter is not 0. Note that there are at most $k C$ such distances, where $C$ is the number of classes seen so far. %% Then, each subvector $\tilde{\textbf{x}}^m_{p}$ is quantized by choosing the closest quantizing state in its corresponding subspace, as described in Equation~(\ref{equa:distance}).
%% we use this quantization to classify each subvector in the corresponding subspace independently from the others by giving the class of the closest quantizing state denoted $c_p$.
The class of the closest average anchor subvector is considered as the decision for the $p$-th subspace. 
Finally, we apply a majority vote over all subspaces to achieve an aggregate decision (c.f.  Algorithm~\ref{Algo:two}). Note that more elaborate strategies can result in higher accuracy but may require more computation during the learning phase as well as memorisation of previously seen examples. %% $\left(c_p\right)_{1\leq p \leq P}$ to classify the unlabelled input signal $\tilde{\textbf{{s}}}^m$ (c.f. Algorithm \ref{Algo:two}).

%% \begin{equation}
%% \left\{\begin{array}{lll} 
%%   k^{\star}(m,p) &= &\displaystyle{\arg\min_k \| \tilde{\textbf{x}}^m_{p} - \textbf{y}_{pk}\|_2} \\
%%   c_p &= &C_p(k^{\star}(m,p))
%%   \end{array}\right.
%% \label{equa:distance}
%% \end{equation}

\begin{algorithm}
\caption{Predicting the Class of a Test Input Signal}
 \textbf{Input}: input signal $\textbf{s}$\\
\begin{algorithmic}
  \STATE Compute the feature vector $\textbf{x}$ associated with $\textbf{s}$
  \STATE Initialize the vote vector $\textbf{v}$ as the \textbf{0 vector} with dimension $C$
\FOR{$p:=1$ to $P$}
\STATE $v_p = \displaystyle{\arg\min_{c} \left[\min_{i}\| \textbf{x}_{p} - \textbf{y}_{c,p,i}\|_2\right]}$
\STATE $\textbf{v}_{v_p} = \textbf{v}_{v_p} + 1$
%% \STATE $c_p=C_p[k^{\star}(m,p)]$
%% \STATE $\textbf{Cls}[c_p]=\textbf{Cls}[c_p] +1$
\ENDFOR
\STATE $\tilde{C} = \displaystyle{\arg\max_c (\textbf{v}_c)}$

\end{algorithmic}
\label{Algo:two}
 \textbf{Output}: class $\tilde{C}$ attributed to $\textbf{s}$ \\
\end{algorithm}

\subsubsection{Data Augmentation}

We use two data augmentation methods to improve the accuracy and robustness during training and classification.

\textbf{Data Augmentation during Training}\\
To improve the accuracy without increasing memory usage, data augmentation is applied to the training dataset. We generate multiple version of each training input signal, and we consider the resulting dataset as an input to train the model.

\textbf{Data Augmentation during Classification}\\
In addition, we propose to obtain multiple predictions for each input signal $\textbf{s}$ using data augmentation~\cite{ciresan3deep}. The idea is to generate multiple versions of the input signal $\textbf{s}$ that we denote $\left(\textbf{s}_r\right)_{1\leq r \leq R}$. We perform a prediction of the class associated with each $\textbf{s}_r$ independently, and then perform a majority vote to obtain the final prediction.

%% In this paper, the two data augmentation methods consist in generating an horizontally flipped version, and shifting the values (translation) of the matrix representing the input signal $\textbf{s}$.    
%% we do a majority vote using the $R$ obtained classes of the $R$ generated signals to finally classify $\tilde{\textbf{s}}^m$ (c.f. Algorithm~\ref{Algo:four}).

\subsubsection{Remarks}

We point out multiple facts about the proposed method:
\begin{enumerate}[label=\alph*)]
\item The learning procedure performs learning one example at a time,
\item The learning procedure is computationally light as it only requires performing of the order of $d$ operations where $d$ is the dimension of feature vectors,
\item The learning procedure has a small memory footprint, as it only stores averages of feature vectors,
\item The learning procedure is such that adding new examples can only increase robustness of the method, so that there is no catastrophic forgetting,
\item During prediction stage, memory usage is of the order of $kCd$ and thus is independent on the number of examples and grows linearly with the number of classes,
\item During prediction stage, computations are of the order of $kCdR$ elementary operations.
\end{enumerate}

From these facts we derive that TILDA is compliant with criteria 1 and 3 defined in the introduction. In the next section, we devise a set of experiments to evaluate the classification accuracy of the proposed method on challenging datasets (criterion 2).

\section{Experiments}
\label{experiments}
In this section we describe the protocol used to test the proposed method and compare its accuracy and memory usage with other incremental learning methods. 
\subsection{Benchmark Protocol}
We propose an incremental learning scenario in which we have streaming data containing new classes/examples. %We test the methods on the extreme case, when the stream data present only one new class at a time.
We test and compare Budget Restricted Incremental Learning (BRIL), Nearest Neighbour search (NN), Nearest Class Mean classifier (NCM), Learn++, incremental Classifier and Representation Learning (iCaRL), and finally the proposed method (TILDA). Learn++ uses Classification And Regression Trees (CART) as weak classifiers.

%According to~\cite{rebuffi-cvpr2017}, BRIL, TILDA, NN, NCM and Learn++ are fixed data representation methods, which means that the feature extractor used is pre-trained and unchanged, only the classifier is changed and adapted to new data. In contrast, iCaRL is a representation learning method, aiming to train and adapt both the feature extractor and the classifier when receiving new data.

We evaluate the different methods using CIFAR10, CIFAR100 and ImageNet ILSVRC 2012~\cite{russakovsky2015imagenet}. We also use $50$ ImageNet classes which have not been used to train the CNN (denoted ImageNet50), and which contains $900$/$100$ training/test images per class. All methods take the same feature vectors extracted from Inception V3~\cite{szegedy2015rethinking} as input. This requires to modify iCaRL method by replacing its CNN with a fully connected network. In the following, and for the iCaRL method, we use a MultiLayer Perceptron (MLP) with one hidden layer containing $1024$ neurons, and output layer containing $C$ neurons, where $C$ is the number of classes. 

The non-incremental learning methods (NI) used are denoted by TMLP and TSVM. TMLP uses transfer learning to compute feature extractors of input data through Inception V3, and then trains a MLP over feature vectors, using the hyperparameters previously described for iCaRL. TSVM method uses Inception V3 to get feature vectors as well, and uses them to train an SVM using Radial Basis Function kernel. 

Data augmentation used in TILDA generates a horizontal flip of the original image, and shifts the pixels of the image by one pixel at a time (to the left, right, top, bottom, and on the four diagonals). Thus we generate $R=10$ images (8 generated by shifting pixels on the image, one generated by horizontal flip and the original one).

\subsection{Results}
 
As a preliminary experiment, we aim to show that replacing the last layers of Inception V3 by the proposed method does not compromise the performances obtained on Imagenet ILSVRC 2012. The 5-top accuracy is $94.4\%$ when we use TILDA with $p=16$ and $k=30$, and $96.5\%$ when we use the last layers of Inception V3 to classify data. The accuracy obtained by TILDA approaches the one obtained by Inception V3, thus our method does not bring a considerable decrease in performances.

The second experiment is performed on CIFAR10/100, ImageNet50 and ImageNet ILSVRC 2012, in which we show the contribution of data augmentation, NCM-inspired classification, and subspace division on classification accuracy. %We the we compare the accuracy of BIRL and TILDA methods and show the contribution of the improvements.
Therefore, we define three methods: TILDA-DA does not use data augmentation and classifies only the original image, TILDA-NCM disregards NCM inspired classification and uses $k$ feature vectors randomly chosen per class, and TILDA-P which is TILDA method with no splitting of vectors. Table~\ref{table:table1} summarises the accuracy of TILDA, TILDA-DA, TILDA-NCM and TILDA-P, when performing one-shot learning (learn one example at a time). We notice that TILDA-DA, TILDA-NCM and TILDA-P reach lower accuracy than TILDA, which confirms that the combination of data augmentation with NCM-inspired classification and supspace division can achieve good performances.

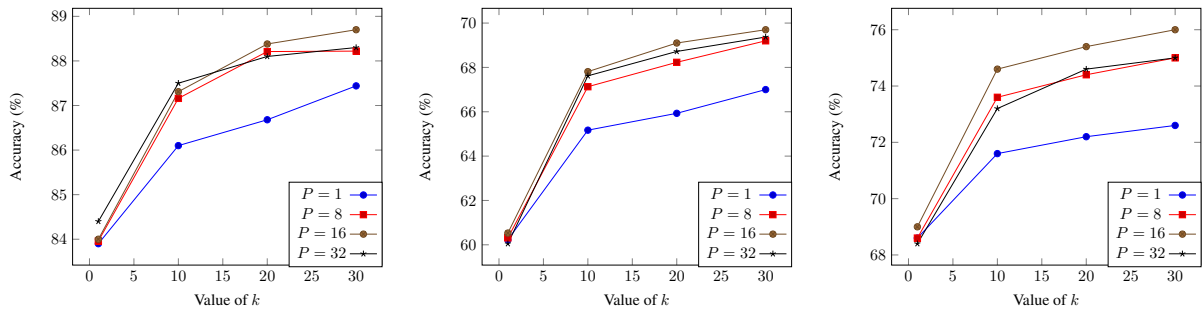
\begin{figure*}
%\multifig% better safe than sorry
\centering
\begin{tabular}{ccc}

\begin{tikzpicture}[scale=0.6]
\begin{axis}[xlabel=Value of $k$,ylabel=Accuracy (\%),
legend style={at={(.699,0.324)},anchor=north west},
legend entries={$P=1$, $P=8$, $P=16$, $P=32$ },
legend plot pos=right]%,enlarge x limits=false,xtick={1,10,100,300},xticklabels={1,10,100,300},point meta=explicit symbolic]

%\addplot[mark=triangle] coordinates {(1,80)(10,80.18)(20,79.80)(30,79.80)(50,79.97)(100,79.94)(200,79.74)(300,80.14)};

\addplot   coordinates {(1,83.9)(10,86.1)(20,86.68)(30,87.44)};
\addplot   coordinates {(1,83.96)(10,87.16)(20,88.21)(30,88.22)};
\addplot   coordinates {(1,84)(10,87.31)(20,88.38)(30,88.7)};
\addplot   coordinates {(1,84.4)(10,87.5)(20,88.1)(30,88.3)};

 \end{axis}
\end{tikzpicture} &
 \begin{tikzpicture}[scale=0.6]
\begin{axis}[xlabel=Value of $k$,ylabel=Accuracy (\%),
legend style={at={(.699,0.324)},anchor=north west},
legend entries={$P=1$, $P=8$, $P=16$, $P=32$},
legend plot pos=right]%,enlarge x limits=false,xtick={1,10,100,300},xticklabels={1,10,100,300},point meta=explicit symbolic]

%\addplot[mark=triangle] coordinates {(1,80)(10,80.18)(20,79.80)(30,79.80)(50,79.97)(100,79.94)(200,79.74)(300,80.14)};

\addplot   coordinates {(1,60.23)(10,65.17)(20,65.93)(30,67)};
\addplot   coordinates {(1,60.31)(10,67.13)(20,68.23)(30,69.2)};
\addplot   coordinates {(1,60.53)(10,67.81)(20,69.1)(30,69.7)};
\addplot   coordinates {(1,60.05)(10,67.62)(20,68.72)(30,69.37)};

 \end{axis}
\end{tikzpicture} &
\begin{tikzpicture}[scale=0.6]
\begin{axis}[xlabel=Value of $k$,ylabel=Accuracy (\%),
legend style={at={(.699,0.324)},anchor=north west},
legend entries={$P=1$, $P=8$, $P=16$, $P=32$},
legend plot pos=right]%,enlarge x limits=false,xtick={1,10,100,300},xticklabels={1,10,100,300},point meta=explicit symbolic]

%\addplot[mark=triangle] coordinates {(1,80)(10,80.18)(20,79.80)(30,79.80)(50,79.97)(100,79.94)(200,79.74)(300,80.14)};

\addplot   coordinates {(1,68.6)(10,71.6)(20,72.2)(30,72.6)};
\addplot   coordinates {(1,68.6)(10,73.6)(20,74.4)(30,75)};
\addplot   coordinates {(1,69)(10,74.6)(20,75.4)(30,76)};
\addplot   coordinates {(1,68.4)(10,73.2)(20,74.6)(30,75)};

 \end{axis}
\end{tikzpicture} \\

%\figcaption{fourth} & \figcaption{fifth} & \figcaption{sixth}
\end{tabular}
\caption{Evolution of the accuracy as a function of $P$ and $k$ for CIFAR10 (left), CIFAR100 (middle) and ImageNet50 (right).}
\label{fig:pk}
\end{figure*}

In the third experiment, we study the effect of both quantization parameters $P$ and $k$ on the accuracy of TILDA (c.f. Figure~\ref{fig:pk}). This experiment demonstrates that TILDA reaches best performances for $P=16$. In the following, we perform experiments using TILDA with $P=16$ and $k=30$.

Note that in order to be fair in comparison with other techniques, we do not perform data-augmentation during training or prediction in TILDA in the upcoming experiments.

\begin{figure*}
%\multifig% better safe than sorry
\centering
\begin{tabular}{ccc}

\begin{tikzpicture}[scale=0.6]
\begin{axis}[xlabel=Number of classes,ylabel=Accuracy (\%),
legend style={at={(0,0.385)},anchor=north west},
legend entries={TILDA-DA, NN, NCM, Learn++,iCaRL },
legend plot pos=right]%,enlarge x limits=false,xtick={1,10,100,300},xticklabels={1,10,100,300},point meta=explicit symbolic]

%\addplot[mark=triangle] coordinates {(1,80)(10,80.18)(20,79.80)(30,79.80)(50,79.97)(100,79.94)(200,79.74)(300,80.14)};

\addplot coordinates {(2,97.9)(3,96)(4,94)(5,91.3)(6,88.5)(7,87.44)(8,86.8)(9,86.7)(10,86.8)};
\addplot coordinates {(1,100)(2,98)(3,96)(4,93)(5,90)(6,88)(7,87)(8,86)(9,85.4)(10,85)};
\addplot coordinates {(1,100)(2,97)(3,95)(4,92)(5,89)(6,86)(7,84)(8,83.6)(9,83.4)(10,83.2)};

\addplot coordinates {(1,100)(2,93)(3,89)(4,85)(5,83)(6,81.66)(7,80.8)(8,80.11)(9,80)(10,79.83)};
\addplot[color=green,mark=*] coordinates {(1,100)(2,92)(3,88)(4,84)(5,81)(6,79)(7,77)(8,75)(9,74)(10,73)};

 \end{axis}
\end{tikzpicture} &
 \begin{tikzpicture}[scale=0.6]
\begin{axis}[xlabel=Number of classes,ylabel=Accuracy (\%),
legend style={at={(.619,1)},anchor=north west},
legend entries={TILDA-DA, NN, NCM, Learn++,iCaRL },
legend plot pos=right]%,enlarge x limits=false,xtick={1,10,100,300},xticklabels={1,10,100,300},point meta=explicit symbolic]

%\addplot[mark=triangle] coordinates {(1,80)(10,80.18)(20,79.80)(30,79.80)(50,79.97)(100,79.94)(200,79.74)(300,80.14)};

\addplot coordinates {(1,100)(10,90)(20,83)(30,78)(40,75)(50,73)(60,71)(70,69)(80,67)(90,66)(100,65.3)};
\addplot coordinates {(1,100)(10,89)(20,81)(30,77)(40,74)(50,70)(60,68)(70,66)(80,63.8)(90,63)(100,62)};
\addplot coordinates {(1,100)(10,89)(20,80.57)(30,75)(40,70.9)(50,68.1)(60,65.4)(70,63)(80,60.6)(90,59)(100,58.25)};

\addplot coordinates {(1,100)(10,75.3)(20,61.85)(30,56.93)(40,51.95)(50,48)(60,42.67)(70,40)(80,37.2)(90,35)(100,33.8)};
\addplot[color=green,mark=*] coordinates {(1,100)(10,90)(20,81)(30,72)(40,64)(50,55)(60,48)(70,43)(80,39)(90,38)(100,37)};
 \end{axis}
\end{tikzpicture} &
\begin{tikzpicture}[scale=0.6]
\begin{axis}[xlabel=Number of classes,ylabel=Accuracy (\%),
legend style={at={(.616,1)},anchor=north west},
legend entries={TILDA-DA, NN, NCM, Learn++,iCaRL},
legend plot pos=right]%,enlarge x limits=false,xtick={1,10,100,300},xticklabels={1,10,100,300},point meta=explicit symbolic]

%\addplot[mark=triangle] coordinates {(1,80)(10,80.18)(20,79.80)(30,79.80)(50,79.97)(100,79.94)(200,79.74)(300,80.14)};

\addplot coordinates {(1,100)(10,95)(20,87)(30,80)(40,76)(50,74)};
\addplot coordinates {(1,100)(10,92.5)(20,82)(30,75)(40,70.9)(50,69.7)};
\addplot coordinates {(1,100)(10,91)(20,79)(30,74)(40,70)(50,67)};
\addplot coordinates {(1,100)(10,83)(20,70)(30,63)(40,57)(50,54.2)};
\addplot [color=green,mark=*] coordinates {(1,100)(10,88)(20,77)(30,72)(40,68)(50,64)};
%% Imagenet 2%%%%%%%%%%%%%%%%%%%%%%%%%%%%%%
%\addplot{(1,)(10,)(20,)(30,)(50,)(100,)(200,)(300,)};
%\addplot coordinates {(1,)(10,)(20,)(30,)(50,)(100,)(200,)(300,)};
%\addplot coordinates {(1,)(10,)(20,)(30,)(50,)(100,)(200,)(300,)};
 \end{axis}
\end{tikzpicture} \\

%\figcaption{fourth} & \figcaption{fifth} & \figcaption{sixth}
\end{tabular}
\caption{Evolution of the accuracy as a function of number of classes for CIFAR10 (left), CIFAR100 (middle) and ImageNet50 (right).}
\label{fig:incremental_classe}
\end{figure*}
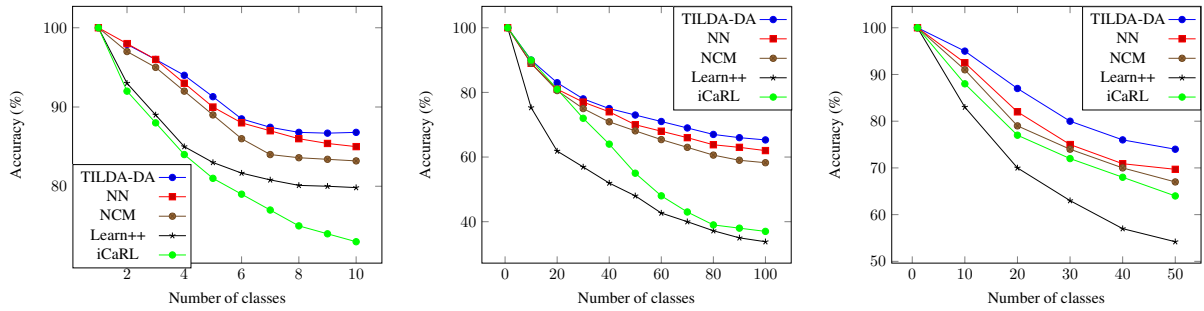

The fourth experiment is stressing the effect of class-incremental learning. We adopt a class-incremental scenario (CI), in which methods are trained over streaming data providing all examples from one class simultaneously, one class at a time. We test and compare TILDA-DA, NCM, Learn++ and iCaRL on CIFAR10/100 and ImageNet50 (c.f. Figure~\ref{fig:incremental_classe}). Learn++ adds one weak classifier each time a novel class is introduced, and iCaRL stores $30$ feature vectors per class. We can see that TILDA-DA outperforms the other methods in this setting.

\begin{figure*}
%\multifig% better safe than sorry
\centering
\begin{tabular}{ccc}
\begin{tikzpicture}[scale=0.6]
\begin{axis}[xlabel=Proportion of database,ylabel=Accuracy (\%),
legend style={at={(.62,.38)},anchor=north west},
legend entries={TILDA-DA, NCM, Learn++, NN,BRIL},
legend plot pos=right]%,enlarge x limits=false,xtick={1,10,100,300},xticklabels={1,10,100,300},point meta=explicit symbolic]

%\addplot[mark=triangle] coordinates {(1,80)(10,80.18)(20,79.80)(30,79.80)(50,79.97)(100,79.94)(200,79.74)(300,80.14)};

\addplot coordinates {(.1,84.8)(.2,85)(.3,85.3)(.4,85.5)(.5,85.7)(.6,85.8)(.7,86)(.8,86.2)(.9,86.4)(1,86.5)};
\addplot coordinates {(.1,82.32)(.2,82.68)(.3,82.76)(.4,82.78)(.5,82.79)(.6,82.83)(.7,82.89)(.8,82.92)(.9,82.93)(1,82.96)};
%\addplot coordinates {(1,100)(10,89)(20,80.57)(30,75)(40,70.9)(50,68.1)(60,65.4)(70,63)(80,60.6)(90,59)(100,58.25)};
\addplot coordinates {((.10,60)(.20,66)(.30,71)(.40,74)(.50,75.7)(.60,77)(.70,78)(.80,78.6)(.90,79.2)(1,79.53)};
\addplot coordinates {(.10,81.32)(.20,82.84)(.40,83.82)(1,85)};
\addplot [color=green,mark=*] coordinates {(0.1,72)(0.2,75.65)(0.3,77.5)(0.4,79.1)(0.6,80.74)(0.7,81.2)(0.8,81.69)(0.9,81.9)(1,82)};

 \end{axis}
\end{tikzpicture} &
 \begin{tikzpicture}[scale=0.6]
\begin{axis}[xlabel=Proportion of database,ylabel=Accuracy (\%),
legend style={at={(.62,.38)},anchor=north west},
legend entries={TILDA-DA, NCM, Learn++, NN,BRIL},
legend plot pos=right]%,enlarge x limits=false,xtick={1,10,100,300},xticklabels={1,10,100,300},point meta=explicit symbolic]

%\addplot[mark=triangle] coordinates {(1,80)(10,80.18)(20,79.80)(30,79.80)(50,79.97)(100,79.94)(200,79.74)(300,80.14)};

\addplot coordinates {(.1,55)(.2,59.8)(.3,61.7)(.4,63)(.5,64)(.6,64.7)(.7,65)(.8,65.15)(.9,65.25)(1,65.3)};
\addplot coordinates {(.1,55)(.2,56.9)(.3,57.2)(.4,57.4)(.5,57.55)(.6,57.7)(.7,57.82)(.8,57.92)(.9,58)(1,58.25)};

\addplot coordinates {((.10,16)(.20,22)(.30,25)(.40,27)(.50,29.28)(.60,31)(.70,32.5)(.80,33)(.90,33.4)(1,33.88)};
\addplot coordinates {(.1,49.66)(.2,53.3)(.3,55)(.4,56.17)(.5,57.2)(.6,57.59)(.7,58.6)(.8,59)(.9,59.9)(1,60.3)};
\addplot [color=green,mark=*] coordinates {(.1,37)(0.2,44)(.3,47)(0.4,50)(.5,52)(0.6,54)(.7,55.7)(0.8,56.4)(.9,56.8)(1,57)};
 \end{axis}
\end{tikzpicture} &
\begin{tikzpicture}[scale=0.6]
\begin{axis}[xlabel=Proportion of dataset,ylabel=Accuracy (\%),
legend style={at={(.62,.38)},anchor=north west},
legend entries={TILDA-DA, NCM, Learn++, NN, BRIL},
legend plot pos=right]%,enlarge x limits=false,xtick={1,10,100,300},xticklabels={1,10,100,300},point meta=explicit symbolic]

%\addplot[mark=triangle] coordinates {(1,80)(10,80.18)(20,79.80)(30,79.80)(50,79.97)(100,79.94)(200,79.74)(300,80.14)};

\addplot coordinates {(.1,70)(.2,72)(.3,73)(.4,73.38)(.5,73.5)(.6,73.6)(.7,73.7)(.8,73.8)(.9,73.9)(1,74)};
\addplot coordinates {(.1,66)(.2,66.34)(.3,66.5)(.4,66.8)(.5,66.9)(.6,67)(.7,67.1)(.8,67.13)(.9,67.17)(1,67.2)};
\addplot coordinates {(.1,32)(.2,33)(.3,36.5)(.4,41)(.5,43.6)(.6,46)(.7,47)(.8,48.5)(.9,49.7)(1,50)};
\addplot coordinates {(.1,62.5)(.2,65)(.3,66.7)(.4,67.5)(.5,68)(.6,68.6)(.7,69)(.8,69.2)(.9,69.4)(1,69.7)};
\addplot [color=green,mark=*] coordinates {(.1,45.7)(.2,50)(.3,54)(.4,57)(.5,59)(.6,61)(.7,63)(.8,64.5)(.9,66)(1,67.4)};
%\addplot coordinates {(1,100)(10,89)(20,80.57)(30,75)(40,70.9)(50,68.1)(60,65.4)(70,63)(80,60.6)(90,59)(100,58.25)};
%\addplot coordinates {((.10,57.44)(.20,71.28)(.30,75.58)(.40,78.03)(.50,80)(.60,81.23)(.70,82.28)(.80,83)(.90,83.28)(1,83.75)};
%\addplot coordinates {(1,100)(10,89)(20,81.64)(30,76.32)(40,70.9)(50,68.1)(60,65.4)(70,63)(80,60.6)(90,59)(100,58.25)};
%\addplot coordinates {(1,100)(2,97.3)(3,94)(4,92)(5,90)(6,88.6)(7,88)(8,87.3)(9,86.4)(10,85.6)};
%% Imagenet 2%%%%%%%%%%%%%%%%%%%%%%%%%%%%%%
%\addplot{(1,)(10,)(20,)(30,)(50,)(100,)(200,)(300,)};
%\addplot coordinates {(1,)(10,)(20,)(30,)(50,)(100,)(200,)(300,)};
%\addplot coordinates {(1,)(10,)(20,)(30,)(50,)(100,)(200,)(300,)};
 \end{axis}
\end{tikzpicture} \\

%\figcaption{fourth} & \figcaption{fifth} & \figcaption{sixth}
\end{tabular}
\caption{Evolution of the accuracy as a function of number of learning examples for CIFAR10 (left), CIFAR100 (middle) and ImageNet50 (right).}
\label{fig:incremental_example}
\end{figure*}
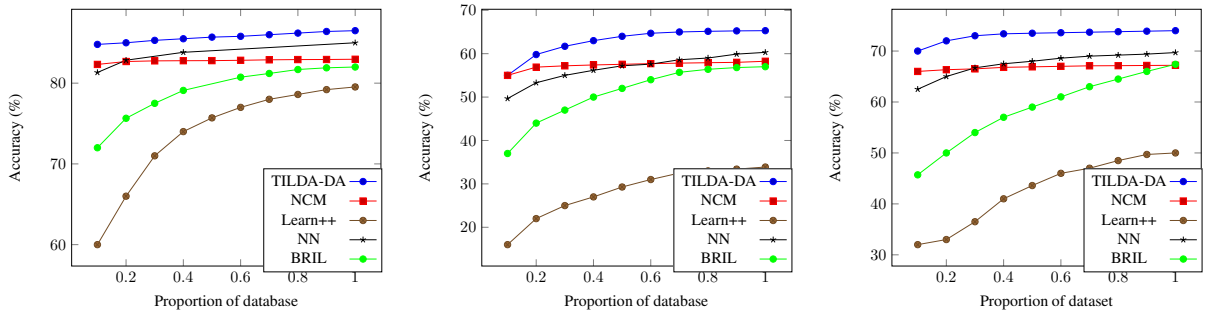

The next experiment illustrates the behaviour of the accuracy when trying to obtain incremental information from new examples of the same class. We adopt an example-incremental scenario (EI), in which we train the method over streaming data providing new examples without introducing new classes. We test and compare TILDA-DA, NCM, NN, Learn++ and BRIL on CIFAR10/100 and ImageNet50. We divide these datasets into $10$ equal size parts, each part containing $5000$ examples ($500/50$ example per class) for CIFAR10/100 and $4500$ examples ($90$ per class) for ImageNet50, and learn one part at a time. Learn++ adds one weak classifier each time a new part is learned. Figure~\ref{fig:incremental_example} shows that all methods handle example-incremental learning and improve their accuracy each time they learn new information provided by new examples. TILDA-DA consistently obtains higher accuracy than Learn++, NCM, NN and BRIL regardless of the quantity of provided data. Note that Learn++ needs large number of examples to perform, and obtains a low accuracy when only few examples are provided.

%%%%% proportion of dataset%%%%%%%

\begin{table*}
%\captionsetup
\centering

    \caption{Accuracy on CIFAR10/100, ImageNet50 and ImageNet ILSVRC 2012. TILDA uses the following parameter:  $P=16$ and $k=30$ . We learn incrementally one example at a time.}
    \centering
    {\renewcommand{\arraystretch}{1.3}%
    \begin{tabular} { | l | l | l | l | l | l |} 
 \hline

    & TILDA  & TILDA-DA & TILDA-NCM & TILDA-P\\
   \hline
 CIFAR100 & $\textbf{69.6\%}$ & $65.3\%$ & $60.7\%$ & $67\%$ \\

   \hline
  CIFAR10 & $\textbf{88.7\%}$ &  $86.6\%$ & $84.11\%$ & $87\%$  \\
  
   \hline
  ImageNet50 & $\textbf{76\%}$ &  $74.4\%$ & $69.2\%$ &  $72\%$  \\
 
   \hline
    ILSVRC 2012 & $\textbf{94.4\%}$ &$91\%$ & $89.6\%$ & $90\%$ \\

   \hline
    \end{tabular}} \quad
    \vspace{.3cm}
  
  \label{table:table1}
  
  \end{table*}

\begin{table*}
 
\centering
    \caption{Comparison of accuracy (Acc) and memory usage (M) relative to full dataset (corresponding to $100\%$) for the different methods. Note that memory usage of Learn++ method represents the size of weak classifiers, and for iCaRL represents the stored feature vectors and the size of the trainable neural network.}
    \centering
    {\renewcommand{\arraystretch}{1.3}%
    \small{\begin{tabular} {| l | l | l | l | l | l | l | l | l |   } 
 
   \cline{2-9}
  \multicolumn{1}{l|}{}  &  \multicolumn{2}{c|}{only CI} & \multicolumn{5}{c|}{both CI and EI} & \multicolumn{1}{c|}{only EI} \\
  \cline{2-9}
   \multicolumn{1}{l|}{}  & Learn++ & iCaRL & TILDA & TILDA-DA & NN & NCM & BRIL & Learn++  \\
   \hline

Acc  (CIFAR100)  & $34\%$  &  $30\%$  & $\textbf{69.6\%}$  & $\textbf{65.3\%}$  & $60.2\%$ &  $58.25\%$ & $57\%$ & $34\%$      \\
    \cline{1-9}
 M  (CIFAR100) & $10.5\%$ & $8\%$ & $6\%$  & $6\%$  & $100\%$ & $\textbf{0.2\%}$ & $6\%$ &  $6.8\%$   \\
   \hline

 Acc  (CIFAR10)  & $79.8\%$  & $41\%$ & $\textbf{88.7\%}$  & $\textbf{86.6\%}$  & $85\%$ &  $83\%$ & $82\%$  & $79.5\%$   \\
   \cline{1-9} 
  M  (CIFAR10) & $0.65\%$  & $2.7\%$ & $0.6\%$  &  $0.6\%$  & $100\%$ & $\textbf{0.02\%}$ & $0.6\%$  & $0.65\%$   \\
  \hline
  Acc  (ImageNet50)  & $54.2\%$  & $64\%$ & $\textbf{76\%}$  & $\textbf{74.4\%}$  & $69.7\%$ &  $67.2\%$ & $67.4\%$  & $50\%$   \\
  \hline
  M  (ImageNet50) & $4.7\%$  & $5.6\%$ & $3.3\%$  &  $3.3\%$  & $100\%$ & $\textbf{0.11\%}$ & $3.3\%$  &  $3\%$  \\
  \hline
  
   \end{tabular}}} \quad
   \vspace{.3cm}
  
 \label{tab:all}
 \end{table*}

Table~\ref{tab:all} presents the different incremental learning methods, obtained accuracies and memory footprints. Learn++ uses either class-incremental scenario (CI) or example-incremental scenario (EI). iCaRL performs learning process using CI. TILDA, NN, NCM, and BRIL use one-shot learning to process one example at a time providing a novel class or additional information, thus they handle both class-incremental and example-incremental at the same time. TILDA outperforms all other incremental learning methods on both accuracy and memory usage.

   The last evaluation we perform aims to compare TILDA with a non incremental learning method such as TMLP and TSVM. To do this, we store and train these methods on the whole dataset. The parameters used for TILDA are $P=16$ and $k=30$ for CIFAR10/100 and ImageNet50. Table~\ref{tab:final} shows that TILDA reaches an accuracy comparable to state-of-art methods.

  As shown by the different evaluation, the TILDA method can at any instant classify data with a good accuracy (c.f. Figure~\ref{fig:incremental_classe} and Figure~\ref{fig:incremental_example}), outperforms other incremental learning methods (c.f. Table~\ref{tab:all}), and approaches state-of-art accuracy (c.f. Table~\ref{tab:final}). Consequently, TILDA fulfils criterion 2.

\begin{table*}
    \caption{Comparison of TILDA with non-incremental learning methods in a non-incremental learning scenario.}
    \centering
    {\renewcommand{\arraystretch}{1.3}%
    \begin{tabular} { | l | l | l | l | l | } 
 \hline

   & TILDA & TILDA-DA & TMLP   & TSVM   \\
   \hline
    Acc (CIFAR100) & $\textbf{69.6\%}$ & $65.16\%$ & $68.6\%$  & $67.6\%$  \\
   \hline
    M (CIFAR100) & $\textbf{6\%}$ & $\textbf{6\%}$ & $100\%$ &  $100\%$  \\
   \hline
    Acc (CIFAR10) & $88.7\%$ & $86.6\%$ & $90\%$ & $89.2\%$ \\
   \hline
    M (CIFAR10) & $\textbf{0.6\%}$ & $\textbf{0.6\%}$ & $100\%$ &  $100\%$ \\
   \hline
    Acc (ImageNet50) & $\textbf{76\%}$ & $74.4\%$ & $75.2\%$  & $75\%$ \\
   \hline
    M (ImageNet50) & $\textbf{3.3\%}$ & $\textbf{3.3\%}$ & $100\%$ &  $100\%$ \\
   \hline
    \end{tabular}} \quad
    \vspace{.3cm}
  
  \label{tab:final}
  \end{table*}

\section{Conclusion}
 In this paper, we have introduced TILDA, a new incremental learning approach inspired by recently proposed methods. TILDA relies on a pre-trained DNN to process data, a projection technique that defines low-dimensional subspaces, NCM inspired classifiers, and data augmentation at both the training and prediction phases. This addresses previous concerns from previous methods of: a) iCaRL as it reaches a good accuracy when stream data contains one class at a time, b) BRIL as it provides a good accuracy comparable to state-of-art method, c) NCM as it uses $k$ anchor vectors instead of one and other methods to increase the accuracy, and d) Learn++ as it still performs well even if steam data does not contains examples of all classes each time. % Two improvements were added to the method and increase the accuracy. An improvement of the training process, by computing and storing the mean feature vectors instead of feature vectors themselves, and an improvement of the testing process, using data augmentation to get numerous images from one image, classify each one and then initiate a majority vote to get the class of the original image.
 Experiments on challenging datasets show that: a) TILDA does not suffer from catastrophic forgetting, in such a way we get the same accuracy and model representation in both incremental learning and offline learning, b) TILDA approaches state-of-art accuracy, c) TILDA uses much less memory usage and gets the same accuracy as nearest neighbour search or even better, d) TILDA still gives a good accuracy even in the case where we have only one class each time, e) and finally, to our knowledge TILDA is the incremental method that reaches the best accuracy. This method is also promising for embedded devices, since it is not necessary to train a DNN or compute extensive operations for learning. 
 
 In future work, we plan to explore further the methods for splitting feature vectors, data augmentation strategies and a weighted majority vote to improve the accuracy. We also plan to propose a hardware architecture of TILDA for incremental learning on chip.
 \label{conclusion}

%%%%%%%%%%%%%%%%%%%%%%%%%%%%%%%%%%%%%%%%%%%%%%%%%%%%%%%%%%%%%%%%%%%%%%%%%%%%%%%%%%%%%%%%%%ùùùùùùùùùùùùùùùùùùùùùùùùùùùùùùùùùùùùùùùùùùùùùùùùùùù%%%%%%%ùùùùùùùùùùùùùùùùùùùùùùùùùùùùùùùùùùùùùùùùùùùùùùùùùùùùùùùù%%%%%%%ùùù
%\newpage
%\newpage
%\pagebreak
\bibliographystyle{aaai}
\bibliography{biblio}

%\bibliography{biblio}
%\bibliographystyle{aaai}

\end{document}